# Biomedical text summarization using Conditional Generative Adversarial Network(CGAN)

Seyyed Vahid Moravvej
Department of Electrical and
Computer Engineering
Isfahan University of Technology
Isfahan 84156-83111, Iran
sa.moravvej@ec.iut.ac.ir

Abdolreza Mirzaei
Department of Electrical and
Computer Engineering
Isfahan University of Technology
Isfahan 84156-83111, Iran
mirzaei@iut.ac.ir

Mehran Safayani
Department of Electrical and
Computer Engineering
Isfahan University of Technology
Isfahan 84156-83111, Iran
safayani@iut.ac.ir

**Abstract**

Text summarization in medicine can help doctors for reducing the time to access important information from countless documents. The paper offers a supervised extractive summarization method based on conditional generative adversarial networks using convolutional neural networks. Unlike previous models, which often use greedy methods to select sentences, we use a new approach for selecting sentences. Moreover, we provide a network for biomedical word embedding, which improves summarization. An essential contribution of the paper is introducing a new loss function for the discriminator, making the discriminator perform better. The proposed model achieves results comparable to the state-of-the-art approaches, as determined by the ROUGE metric. Experiments on the medical dataset show that the proposed method works on average 5% better than the competing models and is more similar to the reference summaries.

## 1. Introduction

Today, the information required by physicians is provided by various sources. These resources include databases of scientific articles, patient medical records storage systems, web-based documents, e-mail reports, and multimedia documents [1, 2]. With the rapid growth of the Internet, the volume of biomedical articles is increasing. PubMed, for example, contains more than 32 million citations to the biomedical articles [3]. Text summarization is an efficient tool for extracting the required information and managing large textual resources.

The purpose of text summarization is to produce a shorter version of the document so that the summary covers all the contents of the document. Summarization approaches are mainly divided into two categories: abstractive and extractive. In extractive summarization, the sentences extracted from the original text are put together and returned as the final summary. In contrast, in abstractive summarization, natural language processing methods are utilized to examine the input text, and a summary containing important contents of the original text is generated. These sentences are very similar to the input text sentences, but they are not the same [4]. From another perspective, summarization can be defined as single-document or multi-document. In single-document summarization, each document is processed independently, and the output is a summary for the given document. In multi-document summarization, multiple documents with the same topic produce the summary [5]. Summarization can be public or query-based. In query-based summarization, the user specifies a query, and system generates a summary which tries to answer the user's query [1]. This paper proposed a public extractive single-document summarization model.

Different methods have been proposed to address the challenges of summarization. These methods mainly use graph [6-8] or machine learning techniques [9-11]. Machine learning methods consider summarization as a classification problem which outputs whether to include a sentence in summary or not. Graph-based methods model the text as a graph and then summarize it by analyzing the nodes. Deep learning has attracted a lot of attention in the fields of natural language processing in different applications, such as machine translation [12, 13], question answering [14, 15], and text classification [16, 17].

Recently, abstractive methods use generative models to generate text. In [18], a generative model based on the reinforcement learning for abstractive summarization is presented. In [19], a method based on auto-encoder and recurrent neural network is proposed. On the other hand, extractive summarization is more practical because it can ensure semantic relevance and grammatical correctness between sentences [20]. The most important challenge in extractive summarization is to select the most valuable sentences of the text so that it covers the important contents of the text.

There are two main components in the extractive summarization techniques: sentence ranking and sentence selection. In most of the proposed methods, including [20-25] for selecting summary sentences, the selected previous sentences are not considered. In other words, they are greedy in choosing the sentences.

The main contributions of this article are as follows:

1- We use a conditional generative adversarial network for summarizing the text with a novel training method.

2- We provide suitable features for use in the text summarization applications.

3- One of the important contributions of the proposed model is the way of scoring the sentences, in which each sentence is scored by considering all the sentences. Another advantage of the model is the production of different summaries for the text. Due to exist of randomness in the generative model, several summaries are generated for each document, in which this article uses the voting system to select proper sentences.

4- In generative adversarial networks, the generator purpose is to produce data close to the real data and the purpose of the discriminator is to distinguish between real and fake data. Training is done under a two-player game. After training, the generator can generate real data. In this paper, many fake summaries are extracted for each document, leading to a new loss function for the discriminator.

The proposed model is evaluated on the medical articles [3]. Articles include biomedicine and health fields and related disciplines such as chemical sciences and bioengineering. Each article has a real summary which is used to evaluate the system. The first proposed method is called GAN-Sum, which uses hand engineering features for summarizing. The second proposed method is E-GAN-Sum, which uses embedding features for this purpose. Experiments show that the proposed method achieves better performance based on the ROUGE metric than the competing algorithms.

The summary of the article is as follows. Section 2 deals with the related works. The backgrounds of the proposed framework are stated in the section 3. The proposed method is introduced in the section 4. Section 5 presents the dataset, evaluation metric, and results. Finally, the article concludes in the section 6.

## 2. Related Works

Recently, many works are presented based on deep learning for text summarization. In this section, we discuss these methods, including those which are compared to our model.

Zhong et al. [21] considered the subject of the document for summarization. In this method, after learning the features of the sentences by an auto-encoder, the importance matrix of document words is created. Then the score of each sentence is calculated using the scores of its words. Finally, the sentences with the highest score are chosen as the summary sentences. Yousefi-Azarand and LenHamey [22] used the same idea. The difference is that instead of using the importance matrix, the cosine similarity between the sentences and the subject is used to score the sentences.

Wu and Hu [20] used reinforcement learning(RL) to summarize the document. They considered coherence as a reward. Their method was the first method which considers the coherence between sentences. The policy function is implemented by a multilayer perceptron (MLP) to select the most valuable sentences for the summary.

Cao et al. [23] suggested recurrent neural networks(RNN) to score sentences. Each sentence is represented as a tree in which the words form the leaves of the tree. Representation of each sentence (tree root) is obtained as a recursive and non-linear process from the leaves of the tree, and the score of each sentence is determined based on its leaves.

Gonzalez et al. [24] proposed a text summarization method based on Siamese neural network (SNN). They used a word and sentence level attention mechanism to rate words and sentences. In this method, the recurrent neural network extracts the features of the word, and the importance of the word in the sentence is estimated using the obtained features. Then the features of the sentence are extracted using the level of attention and the features of the words. The same process is repeated to obtain the features of the document and the reference summary. Finally, using the classifier, the similarity of the summary and the document is obtained. Only the sentence attention level is used to select the summary sentences.

Nallapati et al. [25] proposed a supervised method for single-document summarization using recurrent neural networks. They considered summarization as a classification. First, the embedded vector of words is given to a recurrent neural network, and the hidden vectors of words are extracted. The average of vectors is considered as the vector of sentence features. Finally, logistic regression is used for the binary classification of sentences.

## 3. Backgrounds

### 3.1. Generative Adversarial Network(GAN)

The generative adversarial networks were first proposed by Goodfellow et al. [26]. These networks are one way to create generative models in which two networks are trained simultaneously: a generator which produces fake data and a discriminator which separates real data from the fake data. The generator is trained to deflect the discriminator by producing data which resembles real data.

To learn the generator distribution $p_g$, a mapping function $G(z, \theta_g)$ is used, where $z$ is a noise vector from probability distribution $p_z$, and $\theta_g$ is the generator parameters which must be trained. The function $G(z, \theta_g)$ maps the distribution of $p_z$ to the data space. In addition, the discriminator is equivalent to the function $D(x, \theta_d)$, where $\theta_d$ is the learning parameters, and $x$ is the input data. $D(x)$ indicates the probability that the data $x$ is from the $p_g$ distribution and has a value between zero and one. The output of the discriminator is going to be one for real and zero for fake. Mathematically, $D$ and $G$ play a two-player min-max game with the following value $V(G, D)$:

$$\min_G \max_D V(D, G) = E_{x \sim p_{data}(x)}[log(D(x))] + E_{z \sim p_z(z)}[log(1 - D(G(z)))] \quad (1)$$

where $G$ and $D$ are the generator and discriminator, respectively. $p_{data}(x)$ and $p_z(z)$ represent the real data distribution and input noise, respectively. $E$ is the mathematical expectation.

### 3.2. Conditional Generative Adversarial Network(CGAN)

Generative adversarial networks can be developed to a conditional model, in which the generator and discriminator are conditioned based on additional information. Additional information is any kind of auxiliary information which is given as input to the discriminator and generator [27]. Fig. 1 demonstrates an example of a conditional generative adversarial network. Suppose we want to generate handwritten digits. In this case, the input of the generator includes noise and, the other input is the number that the generator must produce. The output of the generator is the numerical image given at the input as a condition.

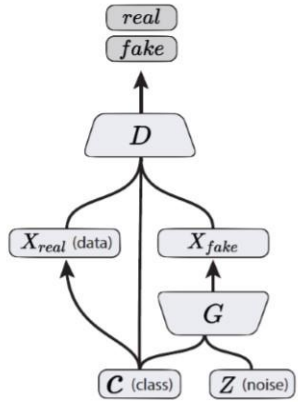

Fig. 1. An example of a conditional generative adversarial network [27].

## 4. The proposed model

The base structure of our model is as a conditional generative adversarial network where the conditional is the feature matrix extracted from the input document. The generator goal is to generate the probability vector of sentences in summary. The proposed model is shown in Fig. 3 and Fig. 4. Fig. 3 shows the generative network of our proposed model. In this network, the document feature matrix is passed through a CNN network, and the output of the CNN and the noise vector are connected and given to the fully connected layer. The output size is equal to the number of summary sentences. Each output has a value between zero and one, which shows the probability of the presence of a sentence in summary. Fig. 4 demonstrates the discriminator network of the proposed model. The structure of this network is similar to the generator. The last layer of the discriminator network is a neuron, which shows the probability of real data.

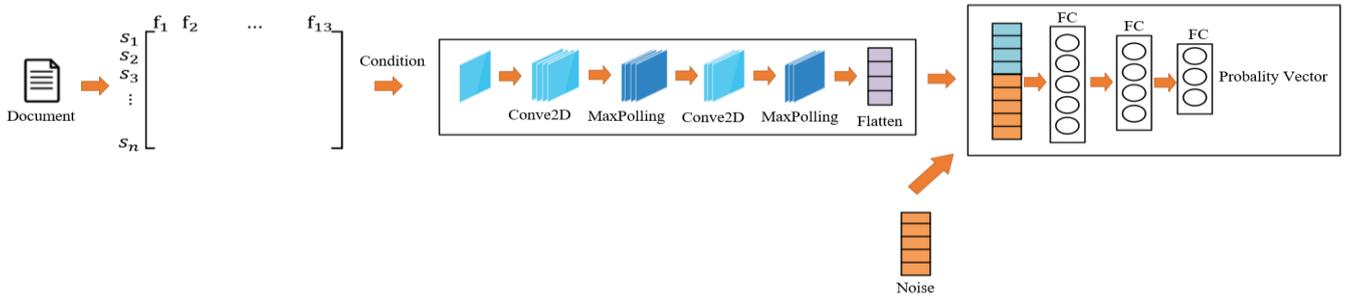

Fig. 2. The generator model.

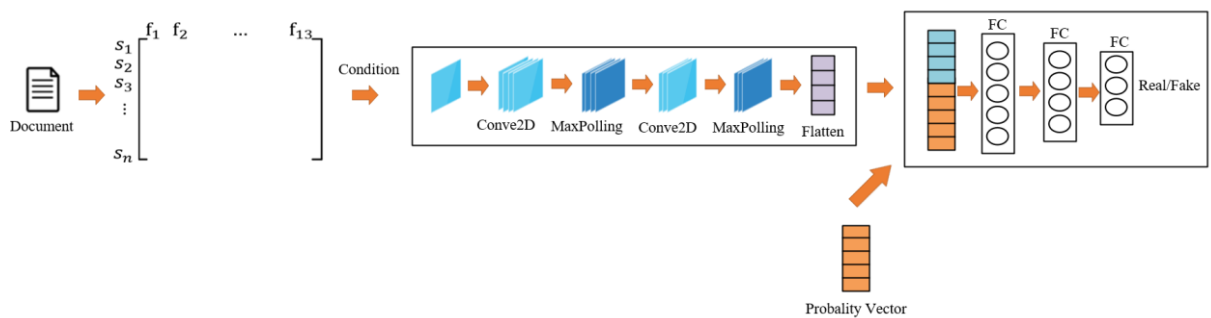

Fig. 3. The discriminator model.

**4.1. Preparing the real and fake data**

In generative adversarial networks, generator output and real data are used for training the discriminator. In this work, to better train the discriminator, synthetic data is also used. Additionally, more than one real data is extracted for each document. Each document actual summaries are text and cannot be used directly to construct real or fake data. For this purpose, a method is needed to extract the real and fake vector using the reference summaries. In this context, a vector for each document is defined where its size is equal to the number of sentences in the document. Each element of this vector is zero or one, which determines the absence or presence of the sentences in the summary. A greedy method is utilized to construct the vectors. The flowchart for making real vectors is shown in Fig. 4. First, a random vector is constructed with $N$ ones, where $N$ is the number of sentences which are in the summary. The vector length is equal to the number of sentences in the document. The sentences with the value of one in this vector are put together to construct a text. Then the obtained text is compared with the reference summary with the Rouge package [28]. The Rouge metric measures the similarity of two texts. After constructing the vector for the document, a randomly chosen one is converted to zero and a randomly chosen zero to one and the Rouge value is recalculated. If the obtained value is more than the previous value (before changing the vector), it will replace the previous vector. This procedure is repeated $Max$ times. Finally, the best vector (the vector with the highest Rogue value) is considered as the document real vector. In this paper, more than one real vector is created for each document. The process of making a fake vector is similar, except that the vector will replace the previous one if the Rogue score is less. It worth mentioning that several identical vectors might be generated in the process. However, they are considered as distinctly generated vectors. In order to have the same length for all documents, the sentences of the document should be equal to a predefined number. For the large documents, a certain of sentences are removed, and for the smaller documents, zero padding is used.

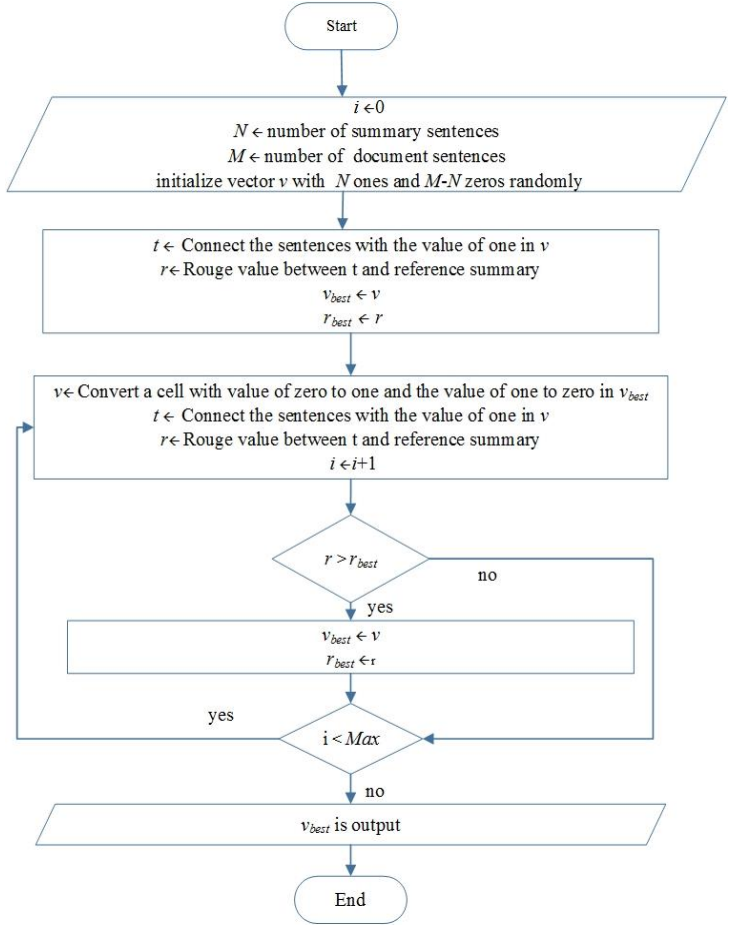

Fig. 4. Flowchart of constructing the real vector.

### 4.2. Loss function

The loss function for the generator is computed as

$$Loss_G = E_{i\sim Data}[E_{z\sim p_{z(z)}}[\log(1 - D(G(z|y_i)))]] \quad (2)$$

where $Data$ is the set of documents at the time of training, $y_i$ is the feature matrix of each document, and $E$ is the mathematical expectation. The loss function for the discriminator network is calculated based on the generator output, real and fake vectors as follows:

$$Loss_D = E_{i\sim Data}[\,E_{z\sim p_{z(z)}}[\log(1 - D(G(z|y_i)|\,y_i))] + E_{k\sim p_{Fake_i}}[\log(1 - D(k|y_i))] \\ + E_{k\sim p_{Real_i}}[\log(D(l|y_i))]] \quad (3)$$

Where $p_{Real_i}$ and $p_{Fake_i}$ represent the distribution of real and fake vectors for the $i$-th document, respectively. Equation 3 causes the discriminator to learn the best and worst summary for each document and forces the generator to produce the best summary. The generator tries to be similar to the sentences given to the discriminator as a real vector.

### 4.3. Summarization

After training, only the generator is used to select the sentences. The selection of sentences in a document is based on voting. First, the feature matrix introduced in the previous sections is extracted for each document and given as the input to the generative network. The probability vector of the sentences is obtained for different noises. After the generator calculates this vector, the sentences which are more likely than average are selected. Finally, the sentences are ranked based on the number of selections, and the sentences with the highest rank are selected as summary sentences.

## 5. Results and discussion

### 5.1. Preprocessing

Pre-processing improves results, reduces computations, increases speed and accuracy. In this article, two preprocessing techniques are used: 1- Stop word removal 2- Stemming.

- **Stop word removal**
  Stop words are words which are of little importance, despite the frequent repetition in the text. At first glance, only relevant words are considered stop words, but many verbs, auxiliary verbs, nouns, adverbs, and adjectives are also considered stop words. It is best to remove these trivial words before extracting the feature.
- **Stemming**
  Another important technique we need to apply is Stemming. Stemming is to get the stem of words by removing suffixes and prefixes so that words with the same stem have the same shape.

### 5.2. Feature extraction

### 5.2.1. Hand engineering features

In this study, we have extracted 13 features for each sentence. Details of the features are listed in Table 1. Some features are at the word level, and some are at the sentence level. All of these features are scaled to [0,1]. To use the features, they must be converted to a matrix, as shown in Fig. 5. The column of this matrix is the extracted features, and the rows of the matrix are the sentences.

Table 1

Extracted features in the proposed model.

| Feature | Description |
| --- | --- |
| Common Word | The number of occurrences (regardless of repetition) N common words in the dataset, divided by the sentence length. |
| Position | The position of the sentence. Supposing there are N sentences in the document, for j the sentence, the position is computed as 1-(j-1)/(N-1). |
| Length | The number of words in the sentence, divided by the length of the largest sentence. |
| Number Raito | The number of digits, divided by the sentence length. |
| Named entity ratio | The number of named entities, divided by the sentence length. |
| Tf/Isf | Term frequency over the sentence, divided by the largest term frequency. |
| Sentence Similarity | The number of occurrences (regardless of repeating) of words in the sentence with the highest Tf/Isf in the sentence divided by the length of the sentence. |
| Stop word | The number of stopwords in the sentence. Supposing there is N number of stop words, for sentence s, the value is computed as 1-N/length(s). |
| None phrase | The number of None phrases, divided by the sentence length. |
| Pos Ratio | A 4-dimensional vector containing the number of nouns, verbs, adjectives, and adverbs. Each cell is divided by the sentence length. |

$$\begin{array}{c} \phantom{s_1}\begin{array}{cccc} f_1 & f_2 & \cdots & f_{13} \end{array} \\ \begin{array}{c} s_1 \\ s_2 \\ s_3 \\ \vdots \\ s_n \end{array} \left[ \phantom{\begin{array}{cccc} f_1 & f_2 & \cdots & f_{13} \end{array}} \right] \end{array}$$

Fig. 5. Features matrix.

### 5.2.2. Feature embedding using deep learning

In another experiment which is performed in order to obtain the effect of word embedding on the model, word embedding as a feature of words is used. Deep neural networks can learn word embedding along with the network parameters. In this work, because the word embedding is used as a condition, and the condition must remain constant, the features are extracted using Skip-Gram [29] before using the proposed model.

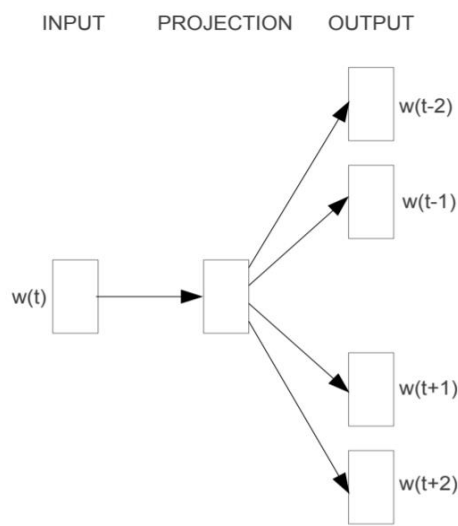

Fig. 6. Skip-Gram architecture [29].

Skip-gram is used to predict the previous and next words of the input word. The architecture of this model is presented in Fig. 6. As we can see, $w(t)$ is the input word, and the purpose is to generate the previous and next words $w(t)$. In this model, there is a hidden layer which calculates the dot product between $w(t)$ and the weight matrix. The output of the hidden layer is transferred to the output layer. The output layer calculates the dot product between the output layer of the hidden layer and the weight matrix of the output layer. Finally, the softmax activation function is used to calculate the probability of words appearing to be in the context of $w(t)$. In this model, w(t) is a one-hot vector with size $v$ (number of words in the dictionary) and the dimension of the weight matrix in the hidden layer is $d \times v$, where $d$ is the size of the embedded vector. After training, the hidden layer matrix is used as the word embedding. In this research, the word embedding dimension is 32.

### 5.3. Dataset

To evaluate the proposed model, 500 medical articles were randomly selected from PubMed Central [3]. The database is divided into three sets of train, validation, and test containing 266, 100, and 134 samples, respectively. The abstract of the article is considered as the reference summary.

### 5.4. Evaluation metric

In this study, the Rouge package presented by Lin [28] was used to evaluate the generated summaries. This metric finds the similarity of the generated summaries of the model with the reference summary and shows it as a score between zero and one. The higher this score, the generated summary has greater similarity with the reference summary. In this article, we used ROUGE-1 and ROUGE-2 scores since they are similar to human judgments [30].

### 5.5. Adjusting the parameters

In this article, two-dimensional convolution is used for the discriminative and generator network. The structure of the discriminative and generator network is the same. The first layer of convolution has 16 filters, and the second layer has 8 filters. The size of the kernel, stride, and padding in both layers are 3, 2, and 1 for both dimensions, respectively. Each convolution layer has a max-pooling layer with dimensions 2 * 2. In order to normalize the data, we pass the data through a batch normalization layer before entering the convolutional neural network. Finally, two fully connected layers have 15 and 20 hidden units, respectively. We used the RELU activation function between layers. The batch size is set to 32. We set the size of noise to 15 for all experiments. In a generative adversarial network, the discriminative network converges faster than the generator, and this causes the generator network to be poorly trained. In order to avoid this problem, in the training process, for every 10 times of generator training, we train the discriminative. We fix the maximum number of sentences to 50. Documents which are longer than this number are truncated. In this study, we considered 100 real vectors and 50 fake vectors for each document.

### 5.6. Results

The proposed summarizer is compared with eight methods. These methods include various approaches, such as statistical, graph-based, and deep learning. Table 2 shows the summary results for the medical dataset at a compression rate 30% for the proposed model and other models. This means that the summary size should be 30% of the document size. When constructing a real vector, each document considers 30 percent of the vector as one. For example, if the vector length (the number of sentences in the document) is 50, 15 cells are one. During training and evaluation, 30% of the sentences which have the highest score in the generator are selected. A brief description of the comparison graph-based models is given below.

**SweSum** [31] is based on the statistical method. This system considers the frequency of the keywords in the document, position of sentences, numerical values, first paragraph tag. SweSum is available for many languages, including English, Persian, Spanish.

**BGSumm** [32] employs the Unified Medical Language System [33](UMLS) to identify the main textual concepts. In this method, concepts are used to build a graph in which the sentences form the graph nodes. The model scores the nodes based on their centrality, and the sentences with the highest score are selected.

**TextRank** [34] is a graph-based summarizer in which sentences form graph nodes. TextRank considers common signs between sentences as a measure of similarity. The obtained similarity values are utilized as the weights of the connections between the nodes. Finally, using graph node rankings, the strongest nodes are selected as summary sentences.

**LexRank** [35] is another graph-based method. It models the sentences as nodes and uses Term Frequency-Inverse Document Frequency (TFIDF) and cosine similarity as a measure of similarity between the sentences and assign weights to the edges. Finally, the sentences with the most centrality are put together as summary sentences.

**TexLexAn** [36] is an open-source summarizer which utilizes features such as keywords to rate sentences.

**CNN** is a basic model which uses only the generator. The purpose of using this model is to investigate the importance of the discriminator network in our model. The training of this basic network is done using the target of each document.

**GAN-Sum** and **E-GAN-Sum** are our proposed model. GAN-Sum uses handcrafted features, and E-GAN-Sum uses features embedding for summarization.

Table 2 shows that deep learning methods performed better than graph-based approaches. In addition, our models work better than other deep learning methods. The reason for the superiority of the proposed method can be due to the way the sentences are scored, where each sentence is scored considering all the sentences. Another reason can be the use of the voting system. Because voting can increase the similarity of the production summary and the reference summary. On the other hand, increasing the number of features and using embedding can improve the proposed model. Fig. 6 shows three sentences extracted and a score assigned to each sentence by the generator. Bold words are shared with the reference summary. As we can see, the sentences with the highest scores contain the most information about the real summary.

Table 2

Evaluation of the proposed model, according to ROUGE.

| Model | R-1 | R-2 |
|---|---|---|
| SweSum [31] | 29.94 | 11.15 |
| BGSumm [32] | 36.19 | 20.02 |
| TextRank [34] | 36.04 | 19.86 |
| LexRank [35] | 35.1 | 19.7 |
| TextLexAn [36] | 35.02 | 19.65 |
| SummaRunner [25] | 37.19 | 21.68 |
| RENS with Coherence [20] | 38.1 | 22.29 |
| SHA-NN [24] | 37.9 | 21.85 |
| CNN | 29.34 | 10.29 |
| GAN-Sum | 40.86 | 24.59 |
| E-GAN-Sum | **43.78** | **26.73** |

| Selected Sentences | Probability of Presence |
|---|---|
| Stancheva said she and other doctors including a psychiatrist diagnosed Burkhart with "**autism, severe anxiety, post-traumatic stress disorder and depression**." | 0.89 |
| **Burkhart**, a 24-year-old **German national**, has been charged with 37 counts of arson following a **string** of 52 **fires in Los Angeles** | 0.86 |
| A medical **doctor** in Vancouver, British Columbia, said Thursday that California arson suspect Harry **Burkhart** suffered from severe mental illness **in 2010**, when she **examined** him as **part of a team** of doctors. | 0.66 |

**Real Summary:** A Canadian doctor says she was part of a team examining Harry Burkhart in 2010, Diagnosis: "autism, severe anxiety, post-traumatic stress disorder and depression", Burkhart is also suspected in a German arson probe, officials say, Prosecutors believe the German national set a string of fires in Los Angeles

Fig. 7. Three sentences extracted by the generator.

## 6. Conclusion and future work

With the increase of textual information in all fields, including medicine, automatic text summarization is popular for researchers. In this paper, we present an extractive summarization approach based on the conditional generative adversarial network for medical texts. To the best of our knowledge, this article is the first proposed system in the field of deep learning for medical datasets. The proposed method consisted of several steps: First, a feature matrix was extracted for each document. Then, the generator predicts the probability of the presence of each sentence in summary. In addition, several real and fake summaries were extracted for each document. We introduced a new loss function to improve the discriminator. In selecting the summary sentences, we used the voting technique, and the sentences that had a higher vote were selected.

The number of features is useful in deep learning methods. In most methods, features are learned by the model. In the proposed system, because the features are used as a condition in the generator and the discriminator, and the condition must be constant for each document, the features for each sentence were first extracted. Of course, these features are considered as primary features, and better features are extracted by the model.

Experimental showed that the proposed models achieved better results than other methods. As future work, we will examine the effect of coherence between sentences which increases readability. One simple approach could be to use coherence to construct a real vector. Therefore, we will try to improve the coherence of the summary sentences as well.